\documentclass[10pt,twocolumn,letterpaper]{article}

\usepackage[pagenumbers]{iccv} 
\usepackage{colortbl}

\newcommand{\bheading}[1]{{\noindent{\textbf{#1}}}}

\definecolor{iccvblue}{rgb}{0.21,0.49,0.74}
\usepackage[pagebackref,breaklinks,colorlinks,allcolors=iccvblue,linkcolor=red]{hyperref}

\usepackage{multirow}  
\usepackage{colortbl}
\usepackage{xcolor}
\usepackage{tikzducks}
\usepackage{float} 
\usepackage{nicematrix}
\newcommand{\algname}{DetAny3D\xspace}
\newcommand{\encodermodule}{2D Aggregator\xspace}
\newcommand{\decodermodule}{3D Interpreter\xspace}
\newcommand{\zeroconv}{ZEM\xspace}
\newcommand{\myparagraph}[1]{\vspace{1pt}\noindent{\bf #1}}

\title{Detect Anything 3D in the Wild}

\author{
Hanxue Zhang$^{1,2\ast}$,
Haoran Jiang$^{1,3\ast}$,
Qingsong Yao$^{4\ast}$,
Yanan Sun$^{1}$,
Renrui Zhang$^{5}$ \\
Hao Zhao$^{6}$,
Hongyang Li$^1$,
Hongzi Zhu$^{2}$,
Zetong Yang$^{1,7}$\\
[2mm]
$^1$~OpenDriveLab at Shanghai AI Laboratory \quad
$^2$~Shanghai Jiao Tong University \quad
$^3$~Fudan University \\ 
$^4$~Stanford University \quad
$^5$~CUHK MMLab \quad
$^6$~Tsinghua University \quad
$^7$~GAC R\&D Center
\\
[1mm]
\normalsize{
\url{https://github.com/OpenDriveLab/DetAny3D}
}}

\begin{document}
\twocolumn[{%
\renewcommand\twocolumn[1][]{#1}
\maketitle
\begin{center}
    \centering
    \vspace{-2.5mm}
    \captionsetup{type=figure}\includegraphics[width=\textwidth]{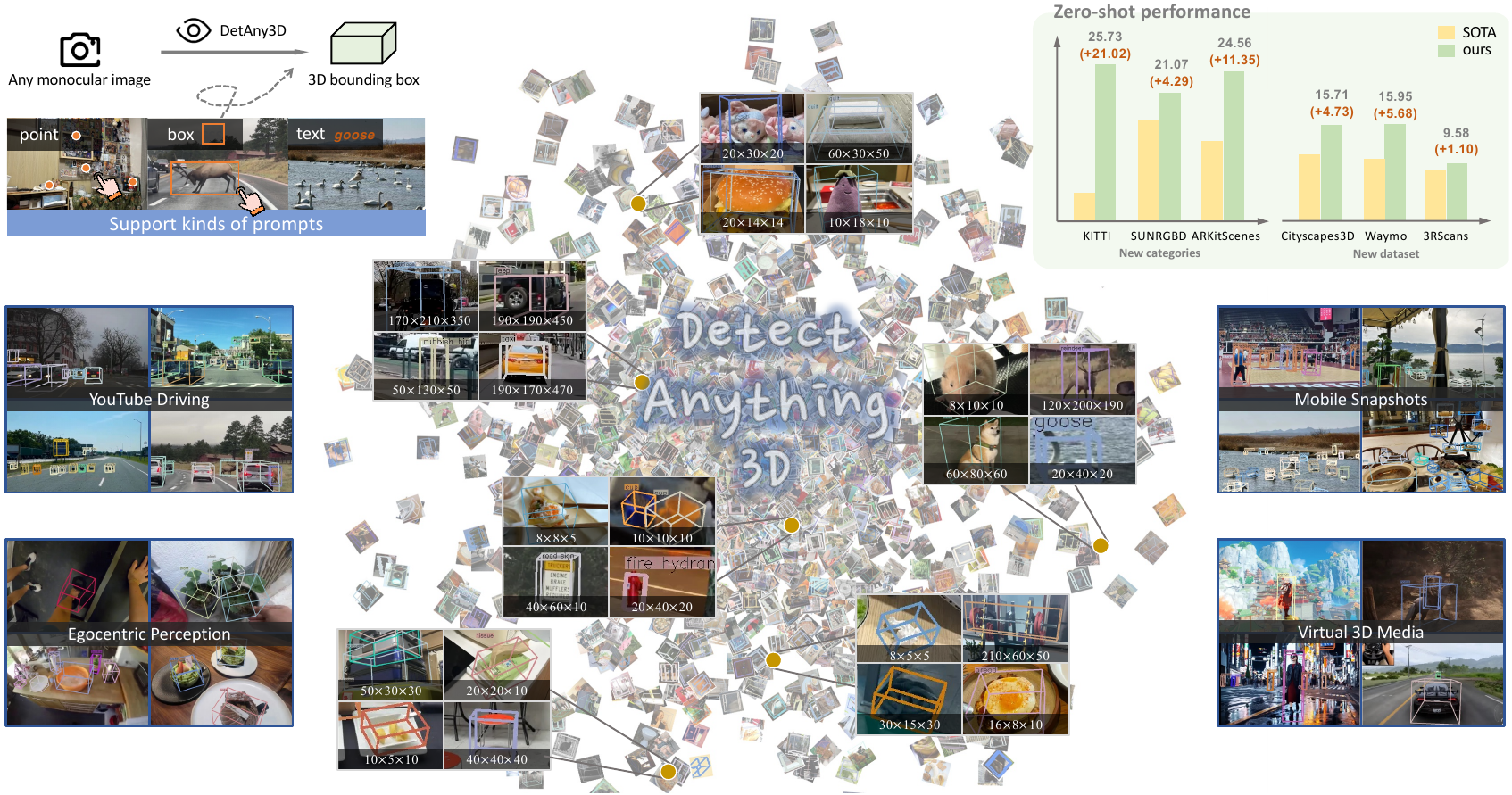} 
\captionof{figure}{
Introducing \textit{DetAny3D}, a promptable 3D detection foundation model capable of detecting any 3D object with arbitrary monocular images in diverse scenes. Our framework enables multi-prompt interaction (\textit{e.g.}, box, point, and text) to deliver open-world 3D detection results $(w \times h \times l$ in {centimeter}) for novel objects across various domains. 
It achieves significant zero-shot generalization, outperforming SOTA by up to 21.02 and 5.68 ${\rm AP_{3D}}$  on novel categories and novel datasets with new camera configurations. 
}
\label{fig:teaser}
\end{center}}]

{\let\thefootnote \relax \footnote{$^\ast$Equal contribution. }}

\begin{abstract}

Despite the success of deep learning in close-set 3D object detection, existing approaches struggle with zero-shot generalization to novel objects and camera configurations. 
We introduce \algname, a promptable 3D detection foundation model capable of detecting any novel object under arbitrary camera configurations using only monocular inputs.
Training a foundation model for 3D detection is fundamentally constrained by the limited availability of annotated 3D data, which motivates \algname to leverage the rich prior knowledge embedded in extensively pre-trained 2D foundation models to compensate for this scarcity. 
To effectively transfer 2D knowledge to 3D, \algname incorporates two core modules: the \encodermodule, which aligns features from different 2D foundation models, and the \decodermodule with Zero-Embedding Mapping, 
which stabilizes early training in 2D-to-3D knowledge transfer.
Experimental results validate the strong generalization of our \algname, which not only achieves state-of-the-art performance on unseen categories and novel camera configurations, but also surpasses most competitors on in-domain data.
\algname sheds light on the potential of the 3D foundation model for diverse applications in real-world scenarios, \eg, rare object detection in autonomous driving, and 
demonstrates promise for further exploration of 3D-centric tasks in open-world settings.
More visualization results can be found at our code repository.

\end{abstract}    
\section{Introduction}
\label{sec:intro}

3D object detection is a fundamental technology for autonomous systems~\cite{li2019gs3d,chen2017multi,chen2016monocular,mao20233d,ma20233d,chen2024end}, robotics~\cite{zhu2014single,biegelbauer2007efficient,wang2024embodiedscan}, and augmented reality~\cite{liu2019edge,park2008multiple}.
3D perception not only enables machines to perceive and interact with the physical world, but also serves as a foundational input for more advanced tasks, such as behavior decision~\cite{fang2020graspnet,asif2018graspnet,casas2021mp3,hu2023planning}, world modeling~\cite{gao2023magicdrive,gao2024magicdrive3d,li2024drivingdiffusion} and 3D scene reconstruction~\cite{nie2020total3dunderstanding,yang2025scenecraft,yao2025cast}.
For practical deployment, a generalizable 3D detector ideally should detect arbitrary objects from easily accessible inputs, such as monocular images, without relying on specific sensor parameters. Such a model would be highly adaptable and reliable for various downstream tasks in diverse and unpredictable environments~\cite{li2019gs3d,chen2017multi,liu2019edge,zhu2014single}.
Also, 
accurate detection results provided by such a detector (\textit{e.g}., generating 3D bounding boxes for even images from the internet) make it a versatile tool, paving the way for scalable 3D systems that leverage Internet-scale data and advance toward open-world scenarios~\cite{gao2023magicdrive,gao2024magicdrive3d,nie2020total3dunderstanding,yang2025scenecraft,li2024drivingdiffusion}.

Previous research, exemplified by Omni3D~\cite{brazil2023omni3d}, has attempted to improve the generalization of the 3D detection system through multi-dataset training~\cite{wang2023uni3detr,kolodiazhnyi2024unidet3d,brazil2023omni3d,li2024unimode}.
However, despite utilizing large datasets to train a unified detector~\cite{brazil2023omni3d,li2024unimode}, these approaches provide limited generalization to novel camera configurations and cannot detect unseen object categories beyond predefined label spaces. 
Therefore, developing a 3D detection foundation model with strong zero-shot generalizability, which is capable of detecting any unseen object under arbitrary camera configurations, remains a crucial and unsolved problem.

While recent advances in 2D foundation models~\cite{kirillov2023segment,oquab2024dinov2,radford2021learning,liu2024grounding} demonstrate remarkable zero-shot capabilities. Segment Anything Model (SAM)~\cite{kirillov2023segment} features a promptable inference mechanism, supporting user-friendly prompts like points and boxes to segment user-specified objects. Their impressive generalization ability stems from training on billions of annotated images.
However, in 3D object detection, the available labeled data is limited to only millions of samples—typically 3-4 orders of magnitude smaller than in 2D images. Such severe data scarcity~\cite{zuo2024towards,yao2024open} poses a fundamental challenge, making it nearly infeasible to train a 3D foundation model from scratch.

In this work, we present \algname, a promptable 3D detection foundation model designed for generalizable 3D object detection using only monocular images (see~\Cref{fig:teaser}). Given the inherent scarcity of 3D annotated data, we achieve strong generalization from two critical perspectives: model architecture and data utilization. The central insight of our approach is to leverage the extensive prior knowledge encoded within two broadly pre-trained 2D foundation models—SAM~\cite{kirillov2023segment} and DINO~\cite{oquab2024dinov2,caron2021emerging}—thus unlocking effective zero-shot 3D detection capabilities with minimal available 3D data.

Specifically, we adopt SAM as our promptable backbone, capitalizing on its versatile and robust object understanding capability derived from large-scale 2D data. Concurrently, we utilize DINO~\cite{oquab2024dinov2} depth-pretrained by UniDepth~\cite{piccinelli2024unidepth}, to offer redundant 3D geometric priors~\cite{yin2023metric3d,bochkovskii2024depth}, which plays a pivotal role for accurate 3D detection in a monocular setting. 
To integrate the complementary features from SAM and DINO more effectively, we propose the \encodermodule, an attention-based mechanism that aligns these features and dynamically optimizes their contributions via learnable gating. \encodermodule fully exploits the strengths of each foundation model.

To further address the challenge of effectively transferring knowledge from 2D to 3D, we introduce the \decodermodule. Central to the \decodermodule is the Zero-Embedding Mapping (\zeroconv) mechanism, which 
ensures stable 2D-to-3D mapping by reducing early-stage interference
and preserving pretrained 2D priors.
By stabilizing the training process across diverse datasets with varying camera parameters, scene complexities, and depth distributions, the \zeroconv mechanism enables progressive zero-shot 3D grounding capabilities, significantly enhancing model generalization.

To leverage as much 3D-related data as possible, we aggregate a diverse range of datasets, including 16 datasets spanning depth with intrinsic data and 3D detection data, refered as DA3D.
Experimental results, using prompts aligned with the baselines, demonstrate three key advantages: 
(1) Generalization to novel classes: achieves 21.0\%, 4.3\%, 11.3\% higher zero-shot ${\rm AP_{3D}}$ than baselines on novel categories on KITTI, SUNRGBD, and ARKitScenes.
(2) Generalization to novel cameras: improves cross-dataset performance by 4.7\%, 5.7\% and 1.1\% ${\rm AP_{3D}}$ compared to baseline methods on zero-shot datasets Cityscapes3D, Waymo and 3RScan.
(3) Performance on in-domain data: surpasses baseline by 1.6\% ${\rm AP_{3D}}$ on Omni3D. Core contributions are summarized in following:
\begin{itemize}
    \item We develop \algname, a promptable 3D detection foundation model capable of detecting any 3D object in real-world scenarios with arbitrary monocular inputs.
     \item \algname introduces \encodermodule to effectively fuse the features from two 2D foundation models SAM and depth-pretrained DINO, which provide pivot shape and 3D geometric priors for various objects, respectively.
     \item In 2D-to-3D knowledge transfer, \algname involves Zero-Embedding Mapping in \decodermodule to 
     achieve stable 2D-to-3D mapping, 
     enabling the model to train robustly across datasets with diverse camera parameters, varying scenes, and different depth distributions.
    \item The experimental results demonstrate significant advantages of \algname, particularly in accurately detecting unseen 3D objects with arbitrary camera parameters in the zero-shot setting, showcasing its potential across a wide range of real-world applications. 
    
\end{itemize}

\section{Related works}
\label{sec:relatedworks}

\subsection{3D Object Detection}
Existing 3D object detection systems have predominantly focused on single-dataset optimization, achieving strong performance on benchmark datasets like KITTI~\cite{geiger2013vision} and nuScenes~\cite{caesar2020nuscenes} through task-specific architectures~\cite{li2024bevformer,zhang2023monodetr,wang2021fcos3d,chen2016monocular,liu2020smoke,dasgupta2016delay,lin2022sparse4d,liang2022bevfusion}. 
While effective in constrained scenarios, these approaches suffer from significant domain gaps when deployed in new contexts, primarily due to their reliance on limited sensor-specific data and closed-set assumptions.
Recent works, exemplified by Omni3D~\cite{brazil2023omni3d}, have demonstrated the potential of multi-dataset training.
Models like Cube R-CNN~\cite{brazil2023omni3d} and UniMODE~\cite{li2024unimode} train a universal monocular 3D detector across multiple datasets, achieving some level of robustness to camera parameters, but are still restricted to predefined classes.
V-MIND~\cite{jhang2025vmind} further addresses the data scarcity challenge by generating pseudo 3D training data from large-scale 2D annotations.
Towards more general detection, OV-Uni3DETR~\cite{wang2024ov} pioneers open-set detection that is able to detect with multimodal inputs, but it is trained separately for indoor and outdoor domains, thereby limiting its overall generalization.
More recently, OVMono3D~\cite{yao2024open} leverages Grounding DINO's~\cite{liu2024grounding} 2D results with a 3D head on unified datasets.
However, it does not fully exploit the priors contained in 2D foundation models, leading to performance constraints tied to the limited 3D data. 
In contrast, our approach fully capitalizes on the knowledge distilled in 2D foundation models while leveraging abundant 3D-related data, thereby enabling the detection of any 3D object from arbitrary monocular inputs.

\subsection{Vision Foundation Models}

Foundation models have demonstrated significant potential across various domains.
For example, language foundation models such as GPT-4~\cite{achiam2023gpt} and DeepSeek~\cite{bi2024deepseek,guo2024deepseek}, trained on massive internet-scale corpora, have achieved impressive capabilities in natural language processing across diverse fields~\cite{touvron2023llama,achiam2023gpt,bi2024deepseek,sima2024drivelm,zhang2024llama,zhang2024mavis}. 
Similarly, foundation models in the vision domain have made remarkable strides~\cite{oquab2024dinov2,kirillov2023segment,liu2024grounding,radford2021learning,li2022grounded,zhang2023personalize,guo2025can}. 
DINOv2~\cite{oquab2024dinov2}, trained on a vast range of curated data from diverse sources, is capable of producing general-purpose visual features that work seamlessly across different image distributions and tasks. 
SAM~\cite{kirillov2023segment} has taken a step further in the vision domain by introducing promptability, enabling models to generalize to novel visual concepts through large-scale data training and continuous model refinement.
In recent years, the development of foundation models in the 3D domain has started to take initial steps~\cite{qi2025gpt4scene,chen2024ll3da,zhu2024unifying,zhu2023ponderv2,zhang2022pointclip,guo2024sam2point}.
Most existing 3D foundation models are often combined with vision-language models (VLMs)~\cite{qi2025gpt4scene,chen2024ll3da,zhu2024unifying,guo2023point}, relying on point clouds as input to help the language models understand 3D~\cite{chen2024ll3da,zhu2024unifying}. 
While these methods are valuable for scene understanding and semantic tasks, they do not directly provide precise 3D detection results. 
Moreover, point cloud inputs significantly restrict the use cases~\cite{Yang_2024_CVPR}, as they are not always accessible in many practical scenarios.
In contrast to these approaches, we aim to develop a foundation model specifically dedicated to 3D detection tasks with the most general inputs, monocular images.
By leveraging the powerful priors from 2D vision foundation models, our approach enables the detection of any 3D object with arbitrary camera configurations, presenting a broad range of practical applications.

\section{Detect Anything 3D in the Wild}
\label{sec:Method}

\subsection{Overview}
\label{sec:overview}
\begin{figure*}[!ht]
 \centering
    \includegraphics[width=\textwidth]{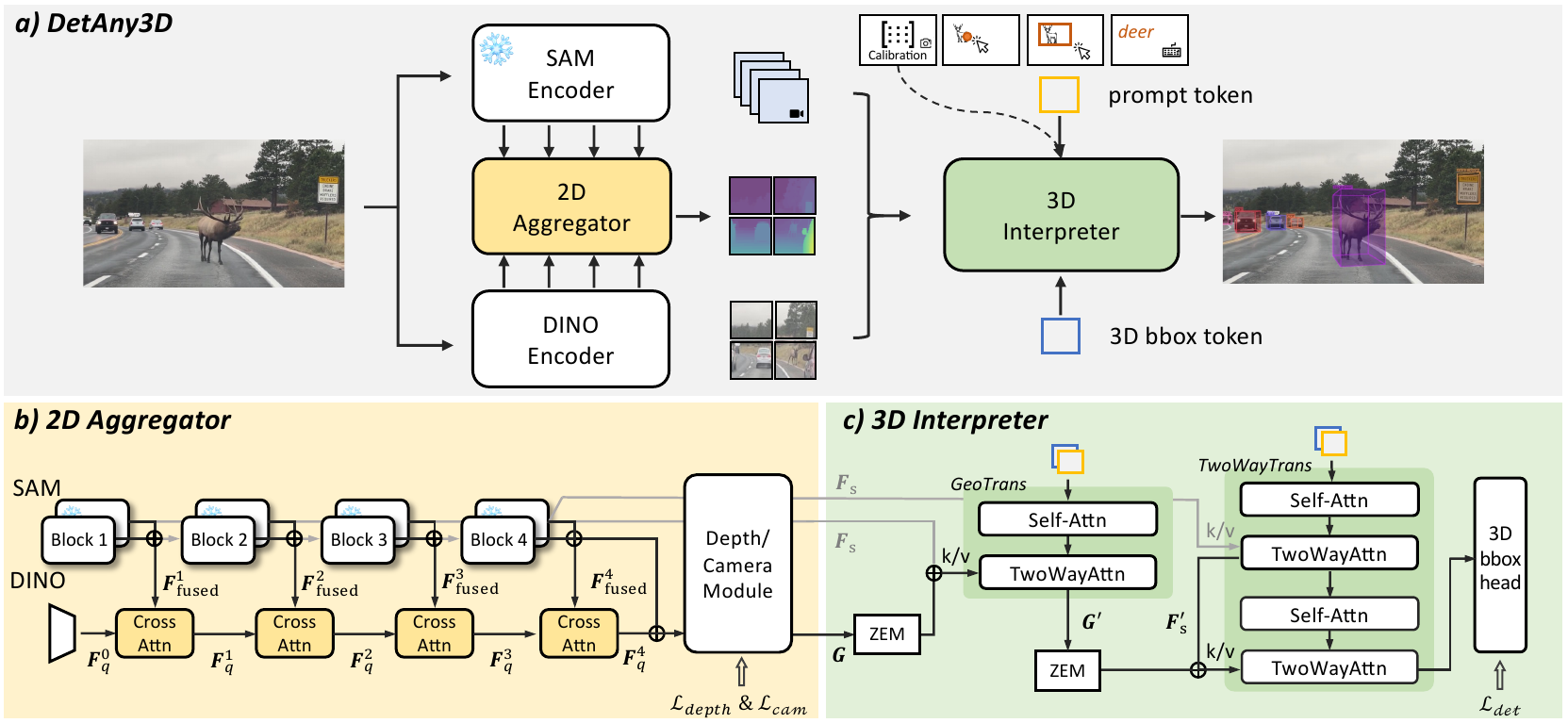}
    \vspace{-6mm}
  \caption{ \textbf{Overview of \algname.} It supports arbitrary monocular images as input and performs 3D object detection driven by prompts—box, point, and text to specify target objects and optional camera calibration to calibrate geometric projections. 
  \algname comprises two key modules: (b) \encodermodule, which employs a hierarchical cross-attention mechanism to dynamically fuse knowledge from SAM and DINO, with a learnable gate controlling each component’s contribution to the geometric embedding; (c) \decodermodule, which introduces a Zero-Embedding Mapping (\zeroconv) strategy based on zero-initialized layers to gradually inject geometric priors, thereby enables zero-shot 3D grounding and avoids catastrophic forgetting during knowledge transfer. }
  \label{fig:pipeline}
  \vspace{-3mm}
\end{figure*}
As illustrated in~\Cref{fig:pipeline}(a), \algname takes a monocular RGB image and prompts (\eg, boxes, points, text, intrinsic) as input. 
The box, point, and text prompts are used to specify objects, while
the intrinsic prompts are optional. 
When not provided, the model predicts the intrinsic parameters and the corresponding 3D detection results. 
If intrinsic are available, the model can leverage them as geometric constraints to mitigate the ill-posed nature of monocular depth estimation and calibrate its detection results.

Specifically, the monocular image is embedded in parallel by two foundational models: SAM~\cite{kirillov2023segment} for low-level pixel information, underpins the entire promptable architecture. And depth-pretrained DINO~\cite{piccinelli2024unidepth, oquab2024dinov2}, 
which provide rich high-level geometric knowledge, excels in depth-related tasks. 
These complementary 2D features are then fused through our proposed \encodermodule (see~\Cref{fig:pipeline}(b)), which hierarchically aligns low-level and high-level information using cross-attention layers. 
The fused features are subsequently passed to the Depth/Camera Module, which extracts the camera and camera-aware depth embedding, collectively referred to as geometric embedding.

The geometric embedding and the 3D bounding box tokens with encoded prompt tokens are then fed into the \decodermodule (see~\Cref{fig:pipeline}(c)), which employs a structure similar to the SAM decoder along with a specialized Zero-Embedding Mapping (\zeroconv) mechanism. 
%
\decodermodule injects 3D geometric features while ensuring stable 2D-to-3D knowledge transfer, enabling progressive 3D grounding across diverse data domains.
%
Finally, the model predicts 3D boxes based on the hidden states of the 3D box tokens. Our \algname is trained on selected seen classes and can detect any unseen classes in a zero-shot manner.

\subsection{\encodermodule}
\label{sec:encoder}

To effectively fuse multiple foundation models
, we propose \encodermodule to aggregate features from SAM and DINO, mitigating potential conflicts between their heterogeneous representations. As illustrated in~\Cref{fig:pipeline}(b), the \encodermodule fuses features from SAM and DINO in a hierarchical manner, progressively integrating spatial and geometric information across four cascaded alignment units.

\bheading{Feature Extraction.} Given an input image, the SAM encoder extracts high-resolution spatial features 
\(
    \mathbf{F}_s \in \mathbb{R}^{H_s \times W_s \times C},
\)
capturing fine-grained details and boundaries. Simultaneously, the DINO encoder outputs geometry-aware embeddings 
\(
    \mathbf{F}_d \in \mathbb{R}^{H_d \times W_d \times C},
\)
which is depth-pretrained by UniDepth~\cite{piccinelli2024unidepth} and provides robust priors for depth and intrinsics. Following the design of ViT Adapter~\cite{chenvision}, we also employ a convolutional structure to produce preliminary image features, denoted as 
\(
    \mathbf{F}_q^0,
\)
serving as the initial query for subsequent attention-based fusion.

\bheading{Hierarchical Fusion.} Each of the four alignment units fuses SAM and DINO features via cross-attention. In the \(i\)-th unit, we first apply learnable gating weights 
\(
    \alpha_i
\)
(initialized to 0.5) to combine the \(i\)-th block of SAM features 
\(
    \mathbf{F}_s^i
\)
and DINO features 
\(
    \mathbf{F}_d^i
\)
as follows:
\begin{equation}
    \mathbf{F}_{\text{fused}}^i 
    = \alpha_i \cdot \mathbf{F}_s^i + (1 - \alpha_i) \cdot \mathbf{F}_d^i.
\end{equation}
We use \(\mathbf{F}_{\text{fused}}^i\) as key and value, while the query feature \(\mathbf{F}_q^{i-1}\) acts as the query in the cross-attention mechanism:
\begin{equation}
    \mathbf{F}_q^i 
    = \texttt{CrossAttn}\bigl(\mathbf{F}_q^{i-1}, \mathbf{F}_{\text{fused}}^i, \mathbf{F}_{\text{fused}}^i\bigr),
\end{equation}
\begin{equation}
    \hat{\mathbf{F}}_{\text{fused}}^i
    = \texttt{Norm}\bigl(\mathbf{F}_{\text{fused}}^i + \mathbf{F}_q^i\bigr).
\end{equation}
This design enables the model to dynamically emphasize SAM’s spatial details or DINO’s semantic and geometric cues at different hierarchy levels while minimizing interference between the two representations.

\bheading{Geometric Embeddings.} The fused features 
\(
    \hat{\mathbf{F}}_{\text{fused}}^i,
\)
\(i \in [1,2,3,4],\) are subsequently processed by the depth and camera modules, following the UniDepth~\cite{piccinelli2024unidepth} architecture. Specifically, these modules predict the camera embedding \(\mathbf{C}\) and camera-aware depth embedding \(\mathbf{D|C}\), referred as the geometric embedding $\mathbf{G} = \{\mathbf{D|C},\, \mathbf{C}\}$. 
These modules provide aligned depth and camera parameters under the monocular depth ill-posed problem.
Further details can be found in the Supplementary material \Cref{sec:supp_unidepth}.

Overall, by progressively aligning multi-scale features and adaptively integrating their contributions, \encodermodule effectively leverages the strengths of both foundation models while minimizing potential conflicts.

\subsection{\decodermodule}
\label{sec:decoder}

The diverse 3D object supervisions across various scenarios, depths, and camera intrinsics introduce challenges to model training.
Our \decodermodule aims to progressively integrate geometric information while ensuring stable 2D-to-3D knowledge transfer.
We introduce Zero-Embedding Mapping (\zeroconv) mechanism, which incrementally infuses 3D geometry into the decoder via zero-initialized layers—without disrupting the original 2D features. As \Cref{fig:pipeline}(c) shows, the \decodermodule comprises three main components: the Two-Way Transformer, the Geometric Transformer, and the 3D bounding box heads. 


\bheading{Two-Way Transformer.}  
Following the SAM design, we first concatenate the 3D bounding box tokens with prompt-related tokens to form the query: 
    \begin{equation}
    \mathbf{Q} = 
\Bigl[
  [\,\mathbf{T}_{\mathrm{3D},1} ; \mathbf{T}_{\mathrm{p},1}],\,
  \cdots,\,
  [\,\mathbf{T}_{\mathrm{3D},N} ; \mathbf{T}_{\mathrm{p},N}]
\Bigr],
\end{equation}
where 
\(\mathbf{T}_{\mathrm{3D},i}\) denotes the 3D bounding box token for the \(i\)-th object,
\(\mathbf{T}_{\mathrm{p},i}\) is the prompt-related token, and
\([\cdot;\,\cdot]\) denotes vector concatenation.
The SAM encoder output \(\mathbf{F}_s\) serves as both key and value for the first Two-Way Transformer layer, yielding:
\begin{equation}
    \mathbf{F}_s' = \texttt{TwoWayTrans}(\mathbf{Q}, \mathbf{F}_s, \mathbf{F}_s).
\end{equation}
The initialized parameters of two-way transformer are copied using pre-trained SAM decoder.

\bheading{Geometric Transformer.}  
We then process the geometric embedding \(\mathbf{G}\) (from the \encodermodule) through the zero-initialized \(1 \times 1\) convolutional layer \(\text{ZEM}\) and add it to \(\mathbf{F}_s\) for use as key and value in the Geometric Transformer:
\begin{equation}
    \mathbf{G}' = \texttt{GeoTrans}\bigl(\mathbf{Q},\, \texttt{ZEM}(\mathbf{G}) + \mathbf{F}_s,\, \texttt{ZEM}(\mathbf{G}) + \mathbf{F}_s\bigr).
\end{equation}
ZEM integrates the geometric embedding and avoids catastrophic forgetting in 2D features.
Next, \(\mathbf{G}'\) is again passed through \(\text{ZEM}\) and combined with \(\mathbf{F}_s'\). This enriched representation is used as key and value in the second Two-Way Transformer layer to generate object features \(\mathbf{O}\) :
\begin{equation}
    \mathbf{O} = \texttt{TwoWayTrans}\bigl(\mathbf{Q}',\, \texttt{ZEM}(\mathbf{G}') + \mathbf{F}_s',\, \texttt{ZEM}(\mathbf{G}') + \mathbf{F}_s'\bigr).
\end{equation}
ZEM also helps stabilize parameter updates in the two-way and geometric transformer training, preventing conflicts arising from diverse 3D object supervision.

\bheading{3D Bounding Box Heads.}  
Finally, \(\mathbf{O}\) is fed into the 3D bounding box heads to calculate the final predictions, which follows typical architectures from standard 3D detection frameworks~\cite{zhang2023monodetr,brazil2023omni3d,wang2021fcos3d}:
    $B_\textrm{3D}(x,\, y,\, z,\, w,\, h,\, l,\, R,\, S\,)$
where \(x, y, z\) specify the 3D box center, \(w, h, l\) are its dimensions, \(R\) is the rotation matrix, and \(S\) is the predicted 3D Intersection over Union (IoU) score.

\subsection{Loss}
\label{sec:loss}

Our loss function comprises three components, the depth loss \(\mathcal{L}_{\text{depth}}\), the camera intrinsic loss \(\mathcal{L}_{\text{cam}}\), and the detection loss \(\mathcal{L}_{\text{det}}\). The overall loss is defined as the sum of these three components.
For depth loss \(\mathcal{L}_{\text{depth}}\), we adopt the commonly used SILog loss~\cite{NIPS2014_7bccfde7,Ummenhofer_2017_CVPR} to supervise depth prediction.
For camera intrinsic loss \(\mathcal{L}_{\text{cam}}\),
we follow the dense camera ray approach~\cite{piccinelli2024unidepth,he2024diffcalib} to represent intrinsics and also employ the SILog loss to measure deviations between predicted and ground-truth parameters.
At last, for detection loss \(\mathcal{L}_{\text{det}}\),
we use the smooth L1 loss~\cite{zhang2023monodetr,li2024unimode,wang2021fcos3d} to regress 3D bounding boxes parameters and predicted IOU scores and the Chamfer loss~\cite{brazil2023omni3d,yao2024open} for rotation matrices. Detailed formulations of these loss functions can be found in the supplementary material~\Cref{sec:supp_loss}.

\subsection{Prompt Interaction}
\label{sec:prompts}


\algname supports point, box, and text prompts to detect 3D box for user-specified objects. To calibrate more precise depth for specific camera, \algname allows users to specify the camera configuration via the intrinsic prompt. 

\bheading{Box and Point Prompts.}
Following SAM's methodology, both box and point prompts are encoded based on their respective positions and embeddings. For the box prompt, two points (top-left and bottom-right corners) are used.
The point prompt is derived by combining the positional encoding of the point and the corresponding embedding.

\bheading{Text Prompts.}
Recent 2D foundation models like Grounding DINO~\cite{liu2024grounding} are able to detect bounding box for the open-vocabulary object specified by users using text prompt. \algname can further generate 3D bouding box using the prediction of Grounding DINO, which enables text as prompts in the zero-shot interface.

\bheading{Intrinsic Prompts.}
Unlike most existing 3D detectors that employ a fixed virtual camera and rely on GT intrinsics to recover the true depth, inspired by Unidepth, we predict intrinsics for camera-aware 3D detection. 
%
When no intrinsic prompt is given, the model infers intrinsics for outputs:
\begin{equation}
    \text{Box}_{3D} 
    = \texttt{3DInterpretor}\bigl(\mathbf{Q},\, \mathbf{\hat{G}},\, \mathbf{F}_s\bigr),
\end{equation}
where  $\mathbf{\hat{G}} = \{\mathbf{D|\hat{C}},\,  \mathbf{\hat{C}}\}$, \(\mathbf{\hat{C}}\) is the predicted camera embedding, and \(\mathbf{D|\hat{C}}\) is the depth embedding conditioned on the predicted camera embedding.
%
When intrinsic prompts are given, the model refines the 3D detection results based on the true intrinsic:
\begin{equation}
    \text{Box}_{3D} 
    = \texttt{3DInterpretor}\bigl(\mathbf{Q},\, \mathbf{G},\, \mathbf{F}_s\bigr),
\end{equation}
where  $\mathbf{G} = \{\mathbf{D|C},\, \mathbf{C}\}$. This boosts performance on both intrinsic prediction and 3D detection since the model continuously predicts and aligns the intrinsic with the 3D detection rather than estimating it solely from input image.

\section{Experiment}
\label{sec:exp}

\begin{table*}[tbp]
    \centering
    \caption{ Zero-shot 3D detection performance comparison on novel categories (left) and novel cameras (right). Results report $\rm{AP_{3D}}$ with different prompt strategies: (1) Cube R-CNN,  (2) \textit{Grounding DINO} outputs (traditional metric / target-aware metric) and (3) \textit{Ground Truth}. Target-aware metric uses per-image existing categories for prompting.}
    \vspace{-3mm}
    \label{Table: combined_zero_shot}
    \resizebox{1.0\linewidth}{!}{
    \begin{tabular}{cccccccc}
    \toprule
    \multirow{2}{*}{Prompt} & \multirow{2}{*}{Method} & \multicolumn{3}{c}{Novel Categories} & \multicolumn{3}{c}{Novel Cameras} \\
    \cmidrule(lr){3-5} \cmidrule(lr){6-8}
     &  & $\rm{AP^{kit }_{3D}}$ & $\rm{AP^{sun }_{3D}}$ & $\rm{AP^{ark }_{3D}}$ & $\rm{AP^{city}_{3D}}$ & $\rm{AP^{wym}_{3D}}$ & $\rm{AP^{3rs }_{3D}}$ \\
    \midrule
    - & Cube R-CNN~\cite{brazil2023omni3d} & - & - & - & 8.22 & 9.43 & - \\
    \midrule
    
    \multirow{3}{*}{Cube R-CNN}
    & OVMono3D~\cite{yao2024open} & - & - & - & 4.97 & 10.89 & - \\
    & \textbf{DetAny3D (ours)} & - & - & - & \textbf{10.33} & \textbf{15.17} & - \\
    & \cellcolor[rgb]{0.9,0.95,1}\textcolor[rgb]{0,0.3,0.6}{\footnotesize $\Delta$} & \cellcolor[rgb]{0.9,0.95,1}\textcolor[rgb]{0,0.3,0.6}{\footnotesize -} & \cellcolor[rgb]{0.9,0.95,1}\textcolor[rgb]{0,0.3,0.6}{\footnotesize -} & \cellcolor[rgb]{0.9,0.95,1}\textcolor[rgb]{0,0.3,0.6}{\footnotesize -} & \cellcolor[rgb]{0.9,0.95,1}\textcolor[rgb]{0,0.3,0.6}{\footnotesize +5.36} & \cellcolor[rgb]{0.9,0.95,1}\textcolor[rgb]{0,0.3,0.6}{\footnotesize +4.28} & \cellcolor[rgb]{0.9,0.95,1}\textcolor[rgb]{0,0.3,0.6}{\footnotesize -} \\
    \midrule
    
    \multirow{3}{*}{Grounding DINO}
    & OVMono3D~\cite{yao2024open} & 4.71 / 4.71 & 4.07 / 16.78 & 13.21 / 13.21 & 5.88 / 10.98 & 9.20 / 10.27 & 0.37 / 8.48 \\
    & \textbf{DetAny3D (ours)} & \textbf{25.73} / \textbf{25.73} & \textbf{7.63} / \textbf{21.07} & \textbf{24.56} / \textbf{24.56} & \textbf{11.05} / \textbf{15.71} & \textbf{15.38} / \textbf{15.95} & \textbf{0.65} / \textbf{9.58} \\
    & \cellcolor[rgb]{0.9,0.95,1}\textcolor[rgb]{0,0.3,0.6}{\footnotesize $\Delta$} & \cellcolor[rgb]{0.9,0.95,1}\textcolor[rgb]{0,0.3,0.6}{\footnotesize +21.02 / +21.02} & \cellcolor[rgb]{0.9,0.95,1}\textcolor[rgb]{0,0.3,0.6}{\footnotesize +3.56 / +4.29} & \cellcolor[rgb]{0.9,0.95,1}\textcolor[rgb]{0,0.3,0.6}{\footnotesize +11.35 / +11.35} & \cellcolor[rgb]{0.9,0.95,1}\textcolor[rgb]{0,0.3,0.6}{\footnotesize +5.17 / +4.73} & \cellcolor[rgb]{0.9,0.95,1}\textcolor[rgb]{0,0.3,0.6}{\footnotesize +6.18 / +5.68} & \cellcolor[rgb]{0.9,0.95,1}\textcolor[rgb]{0,0.3,0.6}{\footnotesize +0.28 / +1.10} \\
    \midrule
    
    \multirow{3}{*}{Ground Truth}
    & OVMono3D~\cite{yao2024open} & 8.44 & 17.16 & 14.12 & 10.06 & 10.23 & 18.05 \\
    & \textbf{DetAny3D (ours)} & \textbf{28.96} & \textbf{39.09} & \textbf{57.72} & \textbf{16.88} & \textbf{15.83} & \textbf{21.36} \\
    & \cellcolor[rgb]{0.9,0.95,1}\textcolor[rgb]{0,0.3,0.6}{\footnotesize $\Delta$} & \cellcolor[rgb]{0.9,0.95,1}\textcolor[rgb]{0,0.3,0.6}{\footnotesize +20.52} & \cellcolor[rgb]{0.9,0.95,1}\textcolor[rgb]{0,0.3,0.6}{\footnotesize +21.93} & \cellcolor[rgb]{0.9,0.95,1}\textcolor[rgb]{0,0.3,0.6}{\footnotesize +43.60} & \cellcolor[rgb]{0.9,0.95,1}\textcolor[rgb]{0,0.3,0.6}{\footnotesize +6.82} & \cellcolor[rgb]{0.9,0.95,1}\textcolor[rgb]{0,0.3,0.6}{\footnotesize +5.60} & \cellcolor[rgb]{0.9,0.95,1}\textcolor[rgb]{0,0.3,0.6}{\footnotesize +3.31} \\
    \bottomrule
    \end{tabular}
    }
        \vspace{-3mm}
\end{table*}

\subsection{Experimental Setup}
\myparagraph{DA3D Benchmark.}
We present DA3D, a unified 3D detection dataset 
that aggregates 16 diverse datasets for 3D detection and depth estimation. Building upon Omni3D's original datasets (Hypersim~\cite{roberts2021hypersim}, ARKitScenes~\cite{baruch1arkitscenes}, Objectron~\cite{ahmadyan2021objectron}, SUNRGBD~\cite{song2015sun}, KITTI~\cite{geiger2013vision}, and nuScenes~\cite{caesar2020nuscenes}), we incorporate additional four outdoor detection datasets (Argoverse2~\cite{wilson2argoverse}, A2D2~\cite{geyer2020a2d2}, Waymo~\cite{sun2020scalability}, Cityscapes3D~\cite{gahlert2020cityscapes}), one indoor detection dataset (3RScan~\cite{wald2019rio}), and five depth and intrinsic datasets (Scannet~\cite{dai2017scannet}, Taskonomy~\cite{zamir2018taskonomy}, DrivingStereo~\cite{yang2019drivingstereo}, Middlebury~\cite{scharstein2002taxonomy}, IBIMS-1~\cite{koch2018evaluation}). All data is standardized with monocular images, camera intrinsics, 3D bounding boxes, and depth maps. Following prior work~\cite{yao2024open}, we select partial categories from KITTI, SUNRGBD, and ARKitScenes as zero-shot test classes. We select Cityscapes3D, Waymo, and 3RScan as our zero-shot datasets with novel camera configurations, where 3RScan also contains novel object categories. Depth supervision from LiDAR, RGB-D, and stereo sensors enhances 75\% of training samples, while intrinsic parameters cover 20 camera configurations across  0.4 million frames (2.5× Omni3D’s scale). Dataset statistics and splits are detailed in Supplementary material~\Cref{sec:supp_da3d}. All data are subject to their respective licenses.

\myparagraph{Baselines.}  
We choose Cube R-CNN~\cite{brazil2023omni3d} and OVMono3D~\cite{yao2024open} as our primary baselines, as their settings align most closely with our experimental protocol:
Cube R-CNN is a benchmark provided by the Omni3D dataset. It is a unified detector capable of performing detection on predefined categories. 
OVMono3D is a recently available open-vocabulary 3D detector on the Omni3D dataset. 
It lifts 2D detection to 3D by connecting the open-vocabulary 2D detector Grounding DINO~\cite{liu2024grounding} with a detection head.

\myparagraph{Metrics.}
We adopt the metrics in the Omni3D benchmark~\cite{brazil2023omni3d}, which is Average Precision (AP). Predictions are matched to ground-truth by measuring their overlap using IoU3D, which computes the intersection-over-union (IoU) of 3D cuboids. The IoU3D thresholds range from $ \tau \in [0.05, 0.10,\dots, 0.50]$.
For experiments using text prompts, we additionally employ target-aware metrics from OVMono3D~\cite{yao2024open}: 
Prompt the detector only with category names present in the per-image annotations instead of providing an exhaustive category list.
This addresses severe naming ambiguity (\eg, "trash can" \vs "rubbish bin") and missing annotation issues prevalent in indoor datasets like 3RScan (see Supplementary material~\Cref{sec:supp_target_aware}.).

\myparagraph{Implementation Details.}
We implement DetAny3D via PyTorch~\cite{paszke2019pytorch}. 
We use the pretrained ViT-L DINOv2~\cite{oquab2024dinov2,piccinelli2024unidepth} and ViT-H SAM~\cite{kirillov2023segment} as our initial models, with SAM serving as the promptable backbone, where the encoder is frozen during training.
All main experiments are conducted using 8 NVIDIA A100 machines with 8 GPUs for each and a batch size of 64. 
The model is trained for 80 epochs, taking approximately 2 weeks to complete. 
The training uses the AdamW~\cite{loshchilov2017decoupled} optimizer with an initial learning rate of 0.0001, adjusted according to the cosine annealing policy~\cite{loshchilov2016sgdr}. 
During box prompt training, we apply a 0.1 positional offset disturbance. 
For point prompt training, points are randomly selected from the mask.
Text prompts are converted into box prompts via Grounding DINO SwinT~\cite{liu2024grounding}.
For fair comparisons, all baseline-related experiments incorporate intrinsic prompts and use aligned prompt inputs.

\begin{table*}[tbp] 
    \centering
    \caption{ 
    In-domain performance comparison between DetAny3D and baselines. The first three columns show results trained only on NuScenes and KITTI, while the next seven columns show results trained on the unified dataset.  Two prompt sources are used: (1) \textit{Cube R-CNN} 2D detections, (2) \textit{Ground Truth}.
    } \label{tab:indomain}
    \vspace{-3mm}
    \resizebox{1\linewidth}{!}{
    \begin{tabular}{l|ccc|ccccccc}
    \toprule
    \multirow{2}{*}{Method} & \multicolumn{3}{c|}{\rm Omni3D\_{OUT}}
    & \multicolumn{7}{c}{\rm Omni3D}\\
   & ${\rm AP^{kit}_{3D}} \uparrow$ 
    & ${\rm AP^{nus}_{3D}} \uparrow$  & ${\rm AP^{out}_{3D}} \uparrow$ 
    & ${\rm AP^{kit}_{3D}} \uparrow$ 
    & ${\rm AP^{nus}_{3D}} \uparrow$ 
    & ${\rm AP^{sun}_{3D}} \uparrow$ 
    & ${\rm AP^{ark}_{3D}} \uparrow$ 
    & ${\rm AP^{obj}_{3D}} \uparrow$ 
    & ${\rm AP^{hyp}_{3D}} \uparrow$ 
    & ${\rm AP_{3D}} \uparrow$ \\
    \midrule

    ImVoxelNet~\cite{rukhovich2022imvoxelnet} &  23.5 & 23.4 & 21.5 & -&-&-&-&-&-& 9.4\\
    SMOKE~\cite{liu2020smoke} &  25.9 & 20.4 &20.0 & -&-&-&-&-&-& 10.4  \\

     OV-Uni3DETR~\cite{wang2023uni3detr} &35.1& 33.0 & 31.6 &  -& -&- &- &- & - & - \\
    Cube R-CNN~\cite{brazil2023omni3d} &  \textbf{36.0} & 32.7 &  31.9   & \textbf{32.50} & 30.06 & 15.33 & 41.73 & 50.84 & 7.48 & 23.26 \\
    
    \midrule
   
    \rowcolor[gray]{0.97}$ \text{OVMono3D~\cite{yao2024open}}_{\textit{w/}~\text{Cube RCNN}}$ & - &-    &-
 & 25.45 &  24.33 & 15.20 & 41.60 & \textbf{58.87} & \textbf{7.75} & 22.98 \\

    \rowcolor[gray]{0.9} \textbf{$\text{DetAny3D (ours)}_{\textit{w/}~\text{Cube RCNN}}$} &  35.8& \textbf{33.9} & \textbf{32.2} & 31.61 & \textbf{30.97}  & \textbf{18.96} &   \textbf{46.13} & 54.42 & 7.17  & \textbf{24.92} \\

   \midrule
    
     \rowcolor[gray]{0.97} $\text{OVMono3D~\cite{yao2024open}}_{\textit{w/}~\text{Ground Truth}}$ &    - &-  &-  & 33.69  &  23.79  & 27.83& 40.85 & 56.64 & 11.99 & 25.32\\
    
    \rowcolor[gray]{0.9} \textbf{$\text{DetAny3D (ours)}_{\textit{w/}~\text{Ground Truth}}$}  & \textbf{38.0}  & \textbf{36.7} & \textbf{35.9} &  \textbf{38.68} & \textbf{37.55} & \textbf{46.14} & \textbf{50.62} &\textbf{56.82} & \textbf{15.98} &  \textbf{34.38 }\\
    \bottomrule
    \end{tabular}
    }
    \vspace{-3mm}
\end{table*}

\subsection{Main Results}
\bheading{Zero-shot Category Performance.}
In this experiment, we use two sources for the prompt input: text prompt processed by Grounding DINO and box prompt from ground-truth 2D bounding box. 
We evaluate our model on KITTI, SUNRGBD, and ARKitScenes datasets with the same zero-shot categories as OVMono3D~\cite{yao2024open}. 
As shown in~\Cref{Table: combined_zero_shot} (left), our DetAny3D demonstrates superior zero-shot adaptation performance compared to the OVMono3D baseline. 
When using Grounding DINO for text prompt input, our method achieves significant improvements of 21.02 ${\rm AP_{3D}}$ on KITTI, 4.29 ${\rm AP_{3D}}$ on SUNRGBD, and 11.35 ${\rm AP_{3D}}$ on ARKitScenes under the target-aware metric. 
When using 2D ground-truth as box prompt input, DetAny3D attains 28.96 ${\rm AP_{3D}}$ on KITTI, 39.09 ${\rm AP_{3D}}$ on SUNRGBD, and 57.72 ${\rm AP_{3D}}$ on ARKitScenes, showing 3.4×, 2.3×, and 4.1× gains over the baseline, respectively. 
This substantial performance gap highlights our method's enhanced ability to generalize to novel object categories.

\bheading{Zero-shot Camera Performance.}
To assess robustness against novel camera parameters, we conduct cross-dataset evaluation as shown in~\Cref{Table: combined_zero_shot} (right). 
For Cityscapes3D and Waymo, We use Cube R-CNN’s 2D detections and ground-truth as box prompt and Grounding DINO processed text prompt for comparison. 
For 3RScan, due to namespace inconsistency with Cube R-CNN’s predefined categories and the presence of novel classes, we only use text prompt and ground-truth box prompts, benchmarking against OVMono3D.
DetAny3D exhibits strong adaptation to unseen camera configurations.
When using Cube R-CNN-aligned prompts, our model achieves $\rm {AP_{3D}}$ scores of 10.33 and 15.17 on Cityscapes3D and Waymo, respectively, surpassing Cube R-CNN by +2.11 and +5.74.
With text prompts, under identical settings as OVMono3D~\cite{yao2024open}, our method improves $\rm{AP_{3D}}$ by +4.73 on Cityscapes3D, +5.68 on Waymo, and +1.1 on 3RScan under \textit{target-aware metrics}. Both models show low scores on conventional metrics for 3RScan due to severe naming ambiguity and missing annotations.
Using 2D ground-truth as box prompts, \algname attains $\rm{AP_{3D}}$ of 16.88, 15.83, and 21.36 across the three datasets, outperforming OVMono3D by +6.82, +5.6, and +3.31, respectively.
These results highlight the effectiveness of our architecture and its potential for real-world applications with arbitrary camera configurations.
\begin{figure}[htbp]
 \centering
    \includegraphics[width=0.48\textwidth]{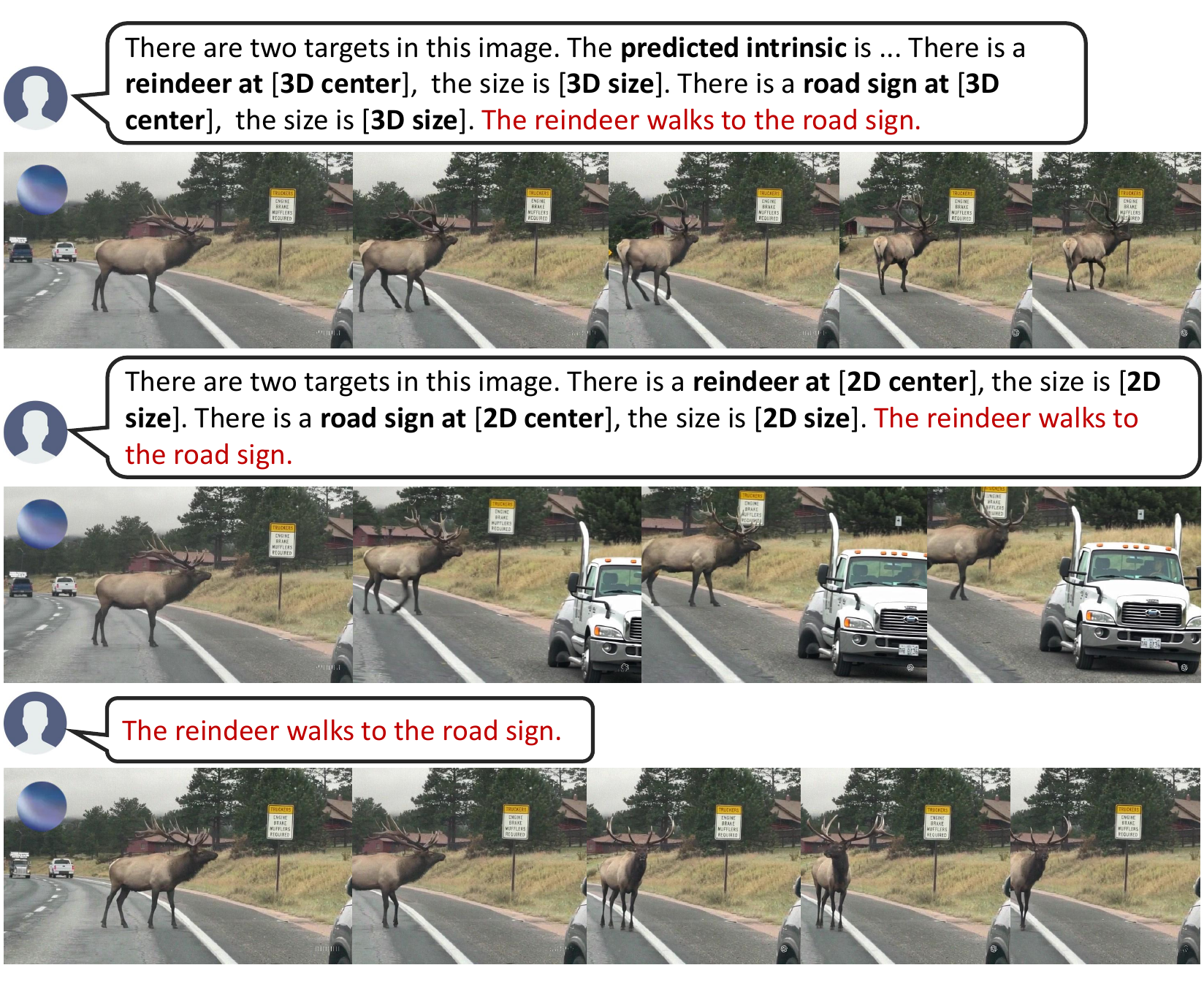}
  \vspace{-6mm}
  \caption{\textbf{Zero-Shot Transfer Video Generation via Sora.} We provide Sora with Internet-sourced images. As shown, when controlled with 3D bounding box, Sora can better capture the scene’s geometric relationships. In contrast, with only controlled by 2D bounding box prompt, Sora respects pixel-level spatial cues but fails to generate accurate geometric offset. 
  }

  \label{fig:sora_downstream}
  \vspace{-6mm}
\end{figure}
\begin{figure*}[htbp]
 \centering
    \includegraphics[width=\textwidth]{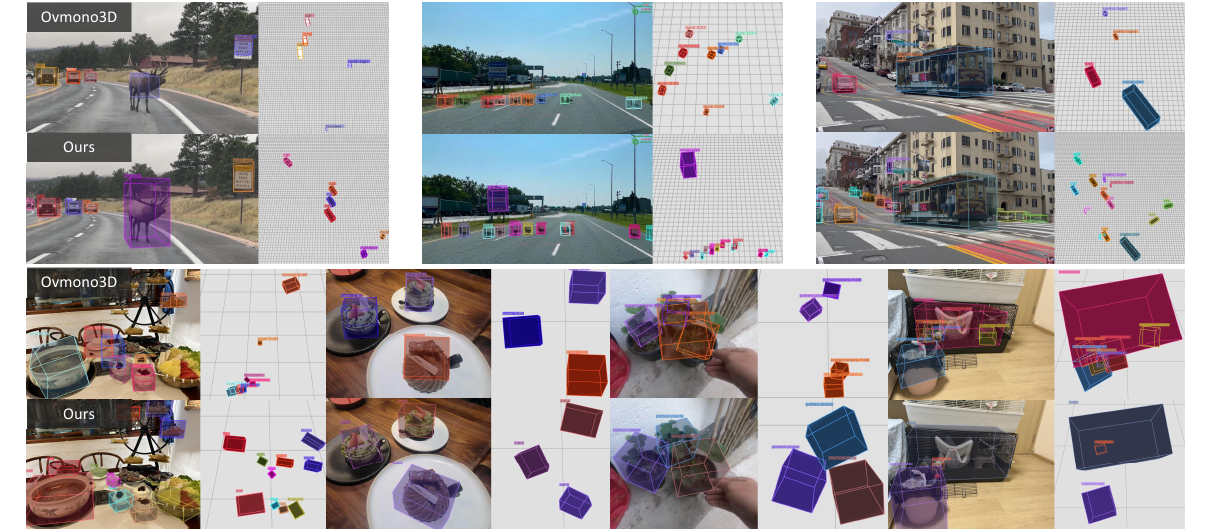}  
  \caption{\textbf{Qualitative Results.} We present qualitative examples from open-world detection. In each pair of images, the top row is produced by OVMono3D, and the bottom row by \algname. For each example, the left sub-figure overlays the projected 3D bounding boxes, while the right sub-figure shows the corresponding bird’s-eye view with 1m$\times$1m grids as the background.}
  \label{fig:visualization}
  \vspace{-2.mm}
\end{figure*}

\bheading{In-domain Performance}
We also evaluate our model's in-domain detection capability using two prompt sources: 2D detections from Cube R-CNN and 2D ground-truth. 
Besides the unified model, we also train \algname on Omni3D\_{out} for comparison. As shown in Table~\ref{tab:indomain}, \algname achieves competitive results with Cube R-CNN when provided with aligned input.
Using GT prompts, \algname outperforms OVMono3D by 9.06 $\rm{AP_{3D}}$, indicating that Cube R-CNN may bottleneck performance, and stronger 2D prompts could further boost results.

\subsection{Possible Applications of \algname}
%
Other than robustly detecting diverse corner cases in real-world tasks such as autonomous driving and embodied perception, \algname's open-world detection results can further serve as inputs for advanced downstream tasks.

\bheading{3D Bounding Box Guided Video Generation.}
We feed \algname outputs into Sora for zero-shot, open-world 3D box guided video generation.
As shown in~\Cref{fig:sora_downstream}, we compare: (i) image + 3D box + text, (ii) image + 2D box + text, and (iii) image + text.
With 3D box constraints, Sora generates videos better aligned with intent.

\subsection{Ablation Studies}
\begin{table}
    \centering
    \caption{\textbf{Ablation study} of DetAny3D. 
The table shows the impact of different design choices on ${\rm AP_{3D}}$ performance. Each component is progressively added. To save resources, ablations are conducted on 10\% of the full training dataset.}

    \label{tab:ablations}
    \small
    \setlength{\tabcolsep}{2.5mm}
    \resizebox{1.0\linewidth}{!}{
    \begin{tabular}{cccc|c}
    \toprule
         Depth\&Cam. & Merge DINO & {2D Agg.} & {\zeroconv} & ${\rm AP_{3D}} \uparrow$  \\
        \midrule

        - & - & - & - & 5.81 \\

         $\surd$ & - & - & - & 10.10 \\
        
        $\surd$ & $\surd$ & - & - & 20.20 \\

        $\surd$ & $\surd$ & $\surd$ & - & 23.21 \\

        $\surd$ & $\surd$ & $\surd$ &  $\surd$ & 25.80  \\
    \bottomrule
    \end{tabular}}
\vspace{-5mm}
\end{table}

As shown in~\Cref{tab:ablations}, we ablate key components of \algname, showing the evolution from a SAM-based baseline to DetAny3D with strong 3D generalization.
The base model extends SAM with 3D box tokens and a 3D head for direct box prediction.
Additional ablations, including backbone and prompt types, are in Supplementary~\Cref{sec:more_ablations}.

\begin{itemize}
    \item \textbf{Effectiveness of Depth \& Camera Modules.}
    Depth map provides denser supervision, while camera configuration intrinsic help mitigate disruptions caused by multiple datasets training.
    Integrating both depth map and camera intrinsic yields improvement in 3D feature extraction and generalization across diverse datasets. 
    
    \item \textbf{Effectiveness of Merging Depth-Pretrained DINO.} Incorporating depth-pretrained DINO yields remarkable improvements, demonstrating that the rich geometric information from DINO effectively compensates for SAM's limited geometric understanding.
    
    \item \textbf{Effectiveness of \encodermodule.} Compared to directly adding the features from two models, the \encodermodule reduces conflicts between different foundation models, further unleashing the performance gains from two foundation model integration.
    
    \item \textbf{Effectiveness of \zeroconv.} \zeroconv mechanism integrates geometric features through zero-initialized layers, which enables stable 2D-to-3D knowledge transfer during training across datasets with diverse camera parameters, scenes, and depth distributions.
    
\end{itemize}

\subsection{Qualitative Results}

We provide qualitative comparisons with OVMono3D.
As shown in~\Cref{fig:visualization}, our model predicts more accurate intrinsics when the camera parameters are unknown and infers more consistent camera poses and 3D detections.

\section{Conclusions}
\label{sec:Conclusions}
We propose \algname, a promptable 3D detection foundation model that can detect arbitrary 3D objects from any monocular image input.
\algname~exhibits significant zero-shot detection capabilities across diverse domains and effective zero-shot transfer across various tasks, highlighting its suitability for real-world deployment in dynamic and unstructured environments. 
Moreover, its flexible and robust detection ability opens the door to gathering large-scale, multi-source data for more 3D perception-guided tasks, paving the way toward open-world systems.

\section*{Acknowledgements}
We sincerely thank Jiazhi Yang, Tianyu Li, Haochen Tian, Jisong Cai, and Li Chen for their invaluable discussions and constructive feedback throughout this project. Their insights and expertise have contributed significantly to the success of this work. We also appreciate the continuous support and encouragement from all the members of OpenDriveLab.
This work is supported by the National Key Research and Development Program of China (2024YFE0210700), the National Natural Science Foundation of China (NSFC) under Grants 62206172 and 62432008, and the Shanghai Artificial Intelligence Laboratory. It is also partially funded by Meituan Inc.

{
    \small
    \bibliographystyle{ieeenat_fullname}
    \bibliography{main}
}
\clearpage
\maketitlesupplementary

\section{DA3D}
\label{sec:supp_da3d}
\begin{figure}[t]
 \centering
    \includegraphics[width=0.48\textwidth]{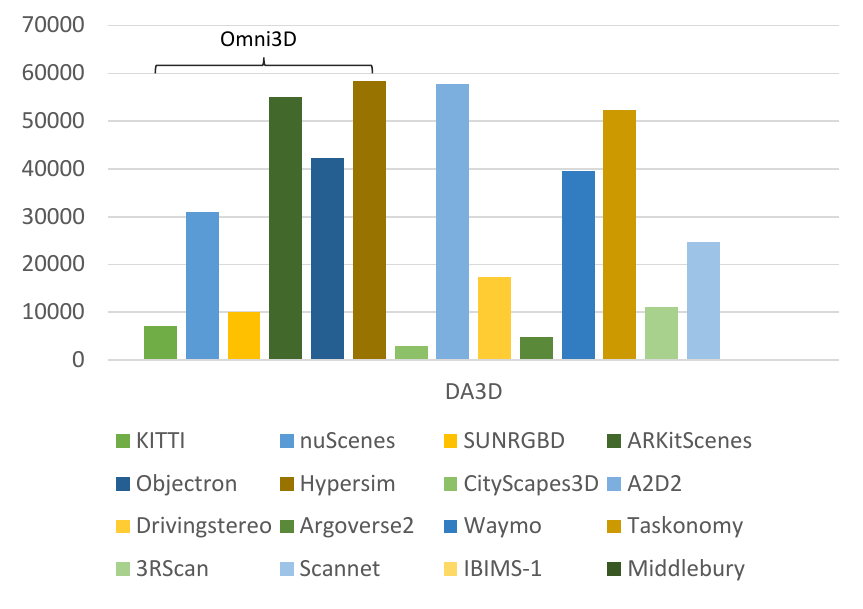}
  \caption{The composition of the DA3D dataset.}
  \label{fig:dataset_compo}
\end{figure}

DA3D is a unified 3D detection dataset, consists of 16 diverse datasets. 
It builds upon six datasets in Omni3D—Hypersim~\cite{roberts2021hypersim}, ARKitScenes~\cite{baruch1arkitscenes}, Objectron~\cite{ahmadyan2021objectron}, SUNRGBD~\cite{song2015sun}, KITTI~\cite{geiger2013vision}, and nuScenes~\cite{caesar2020nuscenes}—while partially incorporating an additional 10 datasets to further enhance the scale, diversity, and generalization capabilities of 3D detection models. As shown in~\Cref{fig:dataset_compo}, DA3D comprises 0.4 million frames (2.5× the scale of Omni3D), spanning 20 distinct camera configurations.

The dataset is standardized with the similar structure to Omni3D~\cite{brazil2023omni3d}, including monocular RGB images, camera intrinsics, 3D bounding boxes, and depth maps. DA3D is designed to test 3D detection models across a wide variety of environments, camera configurations, and object categories, offering a more comprehensive evaluation setting.

\subsection{Dataset Composition}
\begin{figure}[t]
 \centering
    \includegraphics[width=0.48\textwidth]{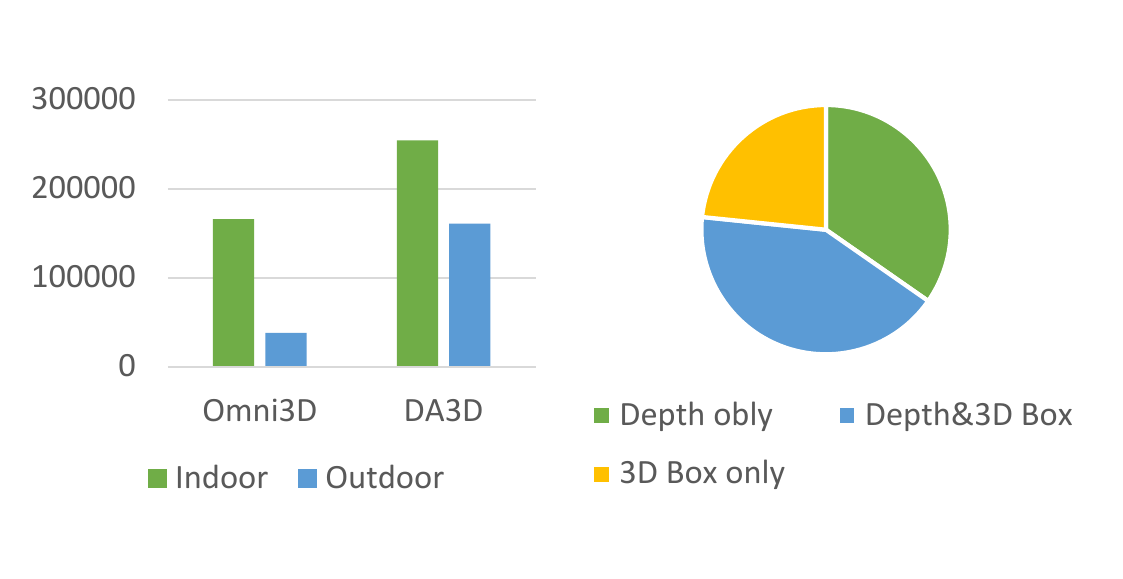}
  \caption{The data distribution of the DA3D dataset. (left): the statistics of indoor and outdoor data. (right): the statistics of data with different supervision categories.}
  \label{fig:dataset_dis}
\end{figure}

We categorize the datasets in DA3D based on two aspects:

\bheading{Indoor \vs. Outdoor.}
As shown in Figure~\ref{fig:dataset_dis} (left), DA3D expands both indoor and outdoor datasets compared to Omni3D. Additionally, the ratio of indoor to outdoor data in DA3D is more balanced than in Omni3D, ensuring a more representative distribution for models trained across diverse environments.

\bheading{Supervision Types.}
We also analyze DA3D in terms of the distribution of supervision types (See \Cref{fig:dataset_dis} (right)):
\begin{itemize}
    \item 35\% data provides only depth supervision.
    \item 23\% data provide only 3D bounding box annotations.
    \item 42\% data contains both depth maps and 3D bounding boxes.
    \item Intrinsic parameters are available for all data.
\end{itemize}

\subsection{Dataset Splits.}
For training and evaluation, we follow the dataset splitting strategy used in prior works~\cite{brazil2023omni3d}. Specifically:
\begin{itemize}
    \item We construct the training set by merging training subsets from the original datasets.
    \item We form the validation set by sampling from the original training data, ensuring balanced representation.
    \item We use the original validation sets of each dataset as the test set, allowing for direct comparison with previous benchmarks.
\end{itemize}

This setup ensures fair evaluation and maintains consistency with existing benchmarks while assessing both in-domain and zero-shot generalization capabilities.

\subsection{Evaluation Setup}
DA3D is designed to evaluate zero-shot generalization in both novel object categories and novel camera configurations. We define two evaluation settings:

\bheading{Zero-Shot Categories.}
Following prior work~\cite{yao2024open}, we select partial categories from KITTI, SUNRGBD, and ARKitScenes as unseen classes for zero-shot testing.

\bheading{Zero-Shot Datasets.}
\begin{itemize}
    \item We use Cityscapes3D, Waymo, and 3RScan as unseen datasets with novel camera configurations.
    \item Cityscapes3D \& Waymo introduce new intrinsics and image styles, challenging models to generalize across different camera setups.
    \item 3RScan not only introduces novel camera setups, but also contains unseen object categories, making it useful for testing both category and camera generalization.
\end{itemize}


\section{Model Details}
\label{sec:model_details}
\subsection{Camera and Depth Module Details}
\label{sec:supp_unidepth}
This section introduces how the camera module and depth module work, predicting intrinsic and camera-aware depth, also related feature. 

\begin{figure}[t]
 \centering
    \includegraphics[width=0.48\textwidth]{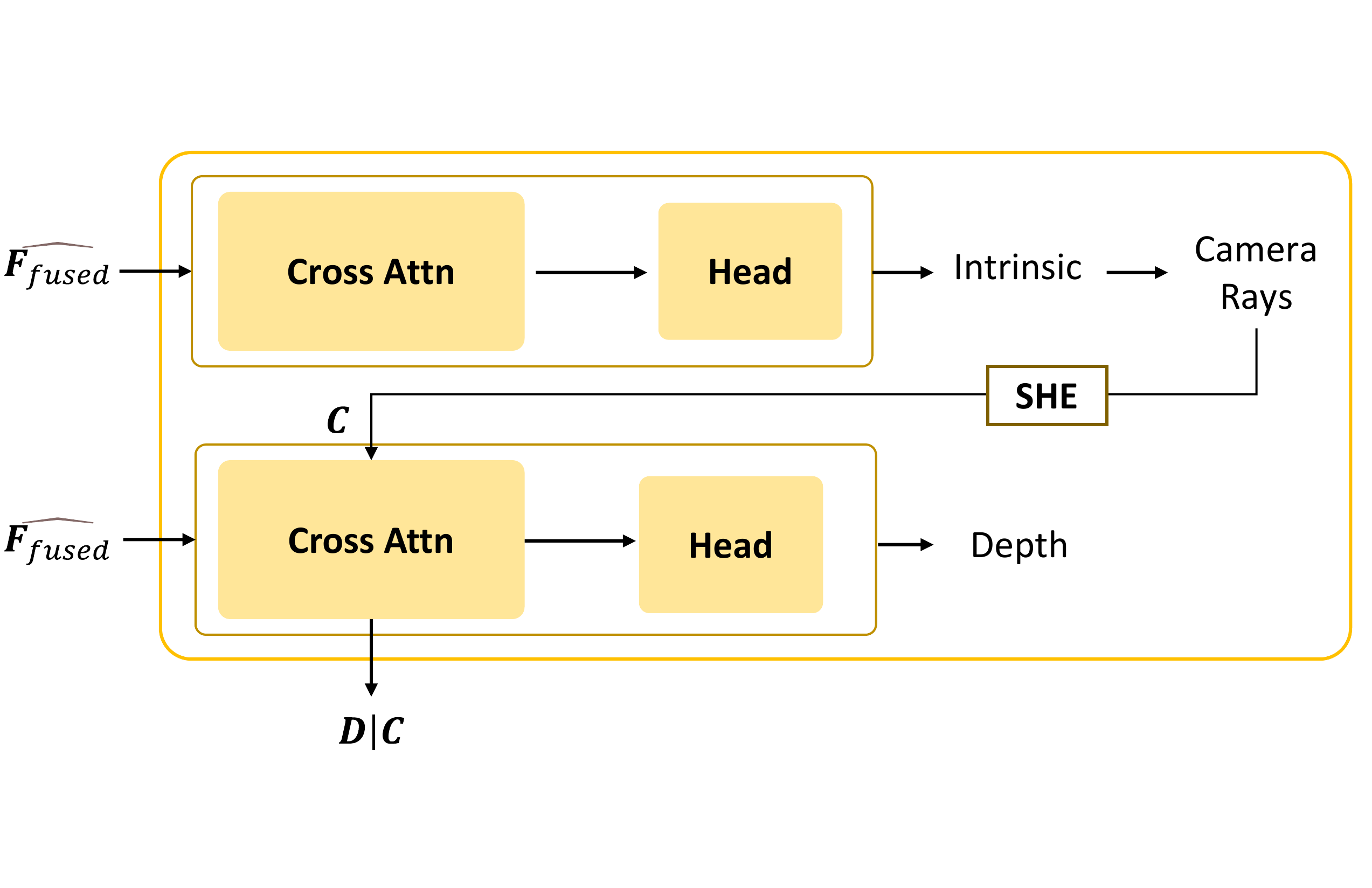}
    \vspace{-3mm}
  \caption{Detailed implementation of camera and depth module from UniDepth.}
  \label{fig:cam_and_depth}
\end{figure}
As show in~\Cref{fig:cam_and_depth}, the fused feature \( \hat{\mathbf{F}}_\text{fused} \) are input into the camera module, which uses a cross-attention mechanism and a to obtain the camera intrinsic parameters. 
These intrinsic parameters are then used to generate camera rays. 
The rays are defined as:

\[
(r_1, r_2, r_3) = \mathbf{K}^{-1} \begin{bmatrix} u \\ v \\ 1 \end{bmatrix}
\]

where \( \mathbf{K} \) is the calibration matrix, \( u \) and \( v \) are the pixel coordinates, and 1 is a vector of ones. In this context, the homogeneous camera rays \( (r_x, r_y) \) are derived from:

\[
\left( \frac{r_1}{r_3}, \frac{r_2}{r_3} \right)
\]

This dense representation of the camera rays undergoes Laplace Spherical Harmonic Encoding (SHE)~\cite{piccinelli2024unidepth} to produce the embeddings \( \mathbf{C} \). 
These embeddings are then passed to the depth module using the cross-attention mechanism.

The depth feature conditioned on the camera embeddings, is computed as:

\[
\mathbf{D}|\mathbf{C} = \texttt{MLP}(\texttt{CrossAttn}(\mathbf{D}, \mathbf{C}))
\]

Subsequently, the depth feature is processed through an upsampling head to predict the final depth map.

\begin{figure}[t]
 \centering
    \includegraphics[width=0.4\textwidth]{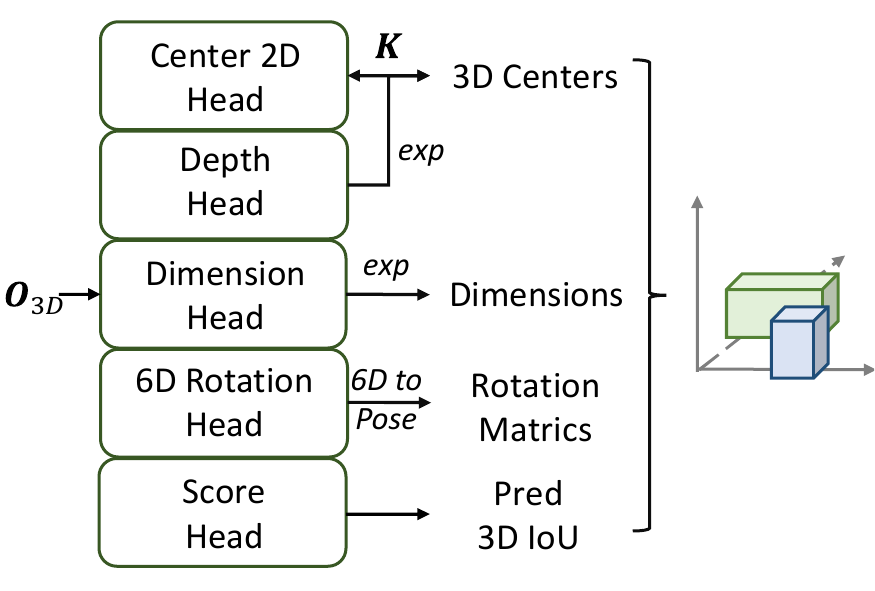}
  \caption{3D Box head details.}
  \label{fig:box_head}
\end{figure}
\subsection{3D Box Head Details}
\label{sec:supp_head}
This section introduces the details of the 3D box head.
After the query $\mathbf{Q}$ passes through the Geometric Transformer and Two-Way Transformer, the model outputs $\mathbf{O}$. $\mathbf{O}$ contains outputs corresponding to both 3D-related hidden states $\mathbf{O}_{3D}$ and prompt hidden states $\mathbf{O}_p$. We extract the 3D-related output $\mathbf{O}_{\text{3D}}$ for further processing.

Subsequently, $\mathbf{O}_{\text{3D}}$ is passed through a series of prediction heads as show in~\Cref{fig:box_head}.

We then transform these predictions into the final 3D bounding box parameters and obtain the 3D bounding box $(x, y, z, w, h, l, R, S)$ for each detected object, where $(x, y, z)$ denotes the 3D center, $(w, h, l)$ represent the dimensions, and $(R, S)$ describe the rotation and predicted 3D IoU score.

\subsection{Loss Details}
\label{sec:supp_loss}

\bheading{Depth Loss.}
The depth module is supervised using the Scale-Invariant Logarithmic (SILog) loss, defined as:
\begin{equation}
\mathcal{L}_{\text{depth}} = \sqrt{ 
  \frac{1}{N} \sum_{i=1}^N {\Delta d_i}^2 - 
  0.15 \cdot \left( \frac{1}{N} \sum_{i=1}^N {\Delta d_i} \right)^2
}
\end{equation}

where $\Delta d_i = \log (d_i^{\text{pred}}) - \log (d_i^{\text{gt}})$, and $N$ is the number of valid depth pixels.

\bheading{Camera Intrinsic Loss.} 
The camera error is computed with the dense camera rays.
For an image with height $H$ and width $W$, the intrinsic loss is formulated as:

\begin{equation}
\mathcal{L}_{\text{cam}} = \sqrt{ 
  \frac{1}{HW} \sum_{i=1}^{HW} {\Delta r_i}^2 - 
1 \cdot \left( \frac{1}{HW} \sum_{i=1}^{HW} {\Delta r_i} \right)^2
}
\end{equation}
where $\Delta r_i = r_i^{\text{pred}} - r_i^{\text{gt}}$.

\bheading{Detection Loss.}
The detection loss consists of three components:
\begin{itemize}
    \item Smooth L1 loss for box regression, covering the prediction of center, depth, and dimensions.

    \item Chamfer loss for rotation matrix prediction, ensuring accurate orientation estimation.

    \item Mean squared error (MSE) loss for 3D IoU score prediction, which optimizes the confidence estimates of detected objects.
\end{itemize}

Combining these terms, the total detection loss is:
\begin{equation}
    \mathcal{L}_{\text{det}} = \mathcal{L}_{\text{box}}
                               + \mathcal{L}_{\text{rot}}
                               + \mathcal{L}_{\text{iou}},
\end{equation}

\section{Target-aware Metrics}
\label{sec:supp_target_aware}
\begin{figure}[t]
 \centering
    \includegraphics[width=0.4\textwidth]{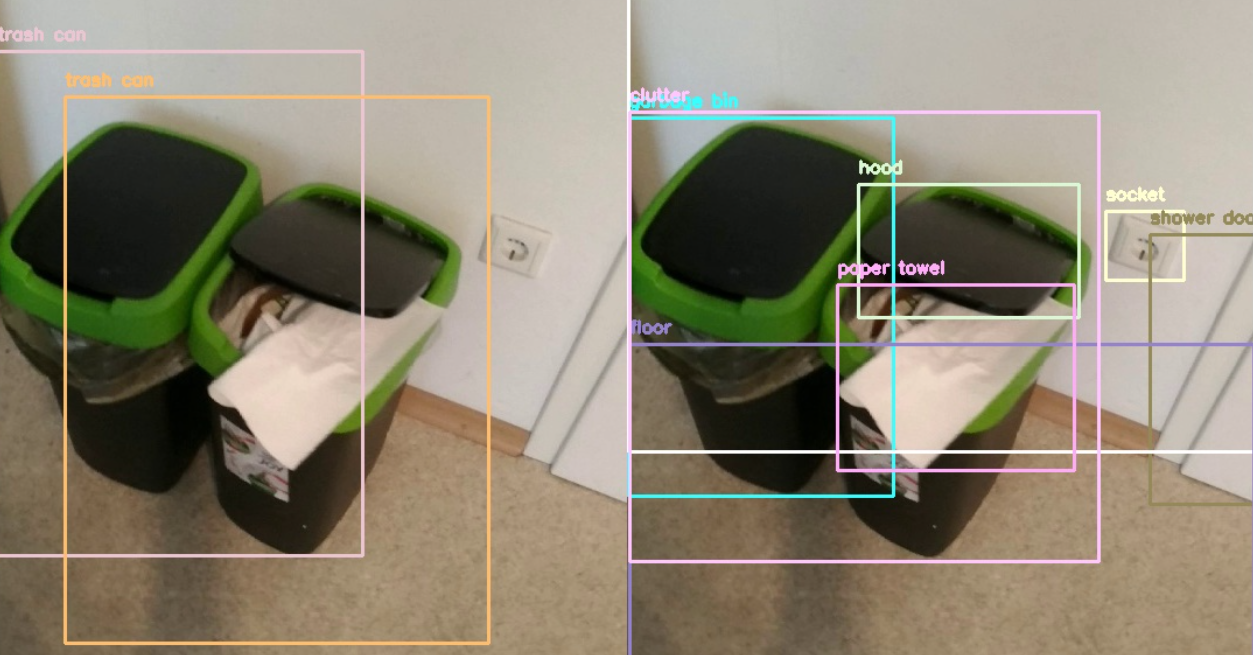}
  \caption{An example on 3RScan. The left image shows the original 3RScan annotations, while the right image presents the detection results from Grounding DINO after feeding in all the 3RScan labels. Severe naming ambiguities (e.g., “trash can” \vs “rubbish bin”) and missing annotations lead to a substantial decrease in the detector’s performance.}
  \label{fig:naming_ambiguity}
\end{figure}

In our work, we evaluate both traditional metrics and the target-aware metrics proposed by OVMono3D~\cite{yao2024open}. 
Under the target-aware paradigm, rather than prompting the model with all possible classes from an entire dataset, we only prompt it with the classes present in the \emph{current} image during inference. 
This is designed to address two key challenges encountered:
\begin{itemize}
    \item \textbf{Missing annotations:} Comprehensive 3D annotation is often impractical or prohibitively expensive, leading to incomplete ground-truth annotations.
    \item \textbf{Naming ambiguity:}  Datasets may label the same objects with inconsistent category names or annotation policies, creating confusion when merging datasets.
\end{itemize}

As illustrated in \Cref{fig:naming_ambiguity}, these issues are especially pronounced in the 3RScan~\cite{wald2019rio} dataset. The left side shows the official 3RScan annotations, while the right side shows detections from Grounding DINO, which are largely misaligned with the dataset’s labeling conventions. Consequently, traditional evaluation metrics may yield misleading or inconsistent results, whereas target-aware metrics help mitigate these mismatches by restricting the evaluated classes to those actually present in the scene.

\section{More Ablation Study}
\label{sec:more_ablations}
\subsection{Various Prompts Performance}
\begin{table}
    \centering
    \caption{Various Prompt Performance.}

    \label{tab:prompts}
    \small
    \setlength{\tabcolsep}{2.5mm}
    \begin{tabular}{c|cccc}
    \toprule
         Prompt Type & Box & Point &  Text   
         \\
         \midrule 
          w/ Intrinsic Prompt &34.38 & 25.19 & 22.31
         \\
        w/o Intrinsic Prompt&  32.16 & 24.0  & 21.02 \\
         
    \bottomrule
    \end{tabular}
\end{table}

In this section, we evaluate different types of prompts, including box prompts, point prompts, and text prompts, both with and without intrinsic prompts. The results on Omni3D are presented in \Cref{tab:prompts}.
Each prompt type demonstrates its effectiveness in guiding 3D detection. Besides, on the zero-shot datasets, we observe that omitting intrinsic prompts leads to a significant performance drop (even approaching zero), which further highlights the critical role of intrinsic prompts for reliable depth calibration in unseen scenarios.

\subsection{Ablation on Different Backbones}
\begin{table}[!tp]
\centering
\caption{Ablation on different backbones. The table reports ${\rm AP_{3D}}$ scores. We verify the effectiveness of SAM and DINO along two dimensions: (1) whether or not we use the pretrained SAM parameters, and (2) whether adopt the pretrained DINO backbone or ConvNeXt for the depth module.}
\begin{tabular}{c|cc}
\toprule
Backbone & w/ SAM &  w/o SAM
    \\

\midrule
 DINO & 25.80 &     19.12    \\ 
ConvNeXt & 23.11 & 18.27
\\
\bottomrule
\end{tabular}
\label{tab:ablation_backbone}
\end{table}

In this section, we investigate our choice of backbone by comparing the use of \emph{SAM} and \emph{DINO} backbones. For DINO, we replace it with ConvNeXt and adopt the same pretraining method proposed by UniDepth. 
For SAM, we examine its effect by removing the SAM-pretrained weights and training from scratch. 
As shown in \Cref{tab:ablation_backbone}, SAM’s pretrained parameters prove crucial for boosting performance. Meanwhile, compared to ConvNeXt, DINO offers richer geometric representations, resulting in stronger 3D detection performance.

\subsection{Ablation on DA3D Dataset}

We ablate the impact of the DA3D dataset in \cref{tab:ab_da3d}. The additional data in DA3D primarily improves generalization to novel cameras, as Omni3D contains only two distinctive intrinsics for outdoor scenes.

\begin{table}[h]
\centering
\caption{Ablation on training datasets. Unless specified, all models are trained on the Omni3D dataset. For the in-domain setting, prompts are provided by Cube R-CNN, while prompts for novel classes and novel datasets are generated by Grounding DINO.}
\resizebox{\linewidth}{!}{
\begin{tabular}{l|c|cc|cc}
\toprule
\textbf{Method} 
& \textbf{In-domain} & \multicolumn{2}{c|}{\textbf{Novel Class}} & \multicolumn{2}{c}{\textbf{Novel Camera}} \\
& ${\rm AP^{omni3d}_{3D}}$ 
& $\rm{AP^{kit}_{3D}}$ & $\rm{AP^{sun}_{3D}}$ 
& $\rm{AP^{city}_{3D}}$ & $\rm{AP^{3rs}_{3D}}$ \\

\midrule
Cube R-CNN &  23.26 & - & - &  8.22 / - & - \\
\midrule
OVMono3D  & 22.98 & 4.71 / 4.71 & 4.07 / 16.78 &  5.88 / 10.98 & 0.37 / 8.48  \\

DetAny3D   & \textbf{24.33} & \textbf{23.75} / \textbf{23.75} & \textbf{7.63} / \textbf{20.87} &   \textbf{8.31} / \textbf{11.68}  & \textbf{0.64} / \textbf{9.56} \\
\midrule
\textcolor{gray!95}{$\rm DetAny3D_{DA3D}$} & \textcolor{gray!95}{24.92} & \textcolor{gray!95}{25.73 / 25.73} & \textcolor{gray!95}{7.63 / 21.07} & \textcolor{gray!95}{11.05 / 15.71} &  \textcolor{gray!95}{0.65 / 9.58}
\\
\bottomrule
\end{tabular}
}
\label{tab:ab_da3d}
\end{table}

\subsection{Ablation on Inference Speed}

We compare the inference speed of DetAny3D with prior methods in \Cref{tab:fps}. DetAny3D runs at 1.5 FPS on a single KITTI image, which is slower than Cube R-CNN (33.3 FPS) and OVMono3D (7.1 FPS). This is a trade-off for stronger generalization across novel categories and cameras, as DetAny3D is designed as a foundation model rather than for real-time deployment.

\begin{table}[h]
\centering
\caption{Inference speed comparison on KITTI.}
\label{tab:fps}
\begin{tabular}{lccc}
\toprule
Method & Cube R-CNN & OVMono3D & DetAny3D \\
\midrule
FPS $\uparrow$ & 33.3 & 7.1 & 1.5 \\
\bottomrule
\end{tabular}
\end{table}

\subsection{Per-category Performance on Novel Classes}
As shown in \Cref{tab:per_category_performance}, we provide a detailed comparison of per-category AP$_\text{3D}$ on novel classes from the KITTI, SUNRGBD, and ARKitScenes datasets between our DetAny3D and the baseline OVMono3D. DetAny3D shows consistent improvements across most categories.

\begin{table}[b]
\centering
\caption{Per-category target-aware AP$_\text{3D}$ comparison on novel classes between DetAny3D and OVMono3D.}

\label{tab:per_category_performance}
\begin{tabular}{l|rr}
\toprule
Category &OVMono3D & DetAny3D \\ \midrule
Board         & 4.83  & 6.02  \\ 
Printer       & 16.23 & 60.22  \\ 
Painting      & 2.80  & 5.11  \\ 
Microwave     & 30.31 & 57.21  \\ 
Tray          & 10.11 & 6.70   \\ 
Podium        & 48.37 & 73.65  \\ 
Cart          & 47.31 & 33.46  \\ 
Tram          & 4.71  & 27.90  \\ 
\midrule
\emph{Easy Categories} & 20.58 & 33.79 \\ 
\midrule
Monitor       & 9.44  & 15.95  \\ 
Bag           & 15.61 & 17.69  \\ 
Dresser       & 29.08 & 41.75  \\ 
Keyboard      & 9.13  & 9.52  \\ 
Drawers       & 43.04 & 40.80  \\ 
Computer      & 7.44  & 12.37  \\ 
Kitchen Pan   & 9.98  & 8.70  \\ 
Potted Plant  & 6.66  & 26.34  \\ 
Tissues       & 12.45 & 12.95  \\ 
Rack          & 10.21 & 9.04  \\ 
Toys          & 5.24  & 16.14  \\ 
Phone         & 3.89  & 4.42 \\ 
Soundsystem   & 13.22 & 6.21 \\ 
Fireplace     & 13.16 & 30.75  \\ 
\midrule
\emph{Hard Categories} & 13.47 & 18.05 \\ 
\midrule
\emph{All Categories} & 16.05 & 23.77 \\ 
\bottomrule
\end{tabular}
\end{table}


\section{Limitations}

\bheading{Text Prompt Process.} Our method leverages open-vocabulary 2D detectors such as Grounding DINO to convert text prompts into 2D box prompts. While effective, this strategy may cause semantic loss, as textual nuances are not directly injected into the 3D detection pipeline. Moreover, 2D detectors are known to perform poorly under heavy occlusion or partial visibility, introducing a domain gap when transferring their outputs to 3D tasks.

\bheading{Inference Efficiency.} Although DetAny3D achieves strong generalization across novel categories and camera settings, its inference speed (1.5 FPS) is significantly slower than existing lightweight 3D detectors. This limits its applicability in latency-sensitive scenarios such as real-time robotics or autonomous driving.

\bheading{Lack of Temporal Modeling.} Our current design operates on single-frame inputs and does not utilize temporal information from video sequences. Incorporating motion cues and enforcing temporal consistency could potentially improve detection accuracy and enable better integration into downstream video-based tasks, such as video knowledge distillation and temporal grounding.

\section{Licenses and Privacy}
All data used in this work are obtained from publicly available datasets and are subject to their respective licenses.

\end{document}